# Hierarchical Vision Transformer Enhanced by Graph Convolutional Network for Image Classification

Haibin Jiao[1]

[1]Independent Researcher
jiaohaibin@alu.ruc.edu.cn

**Abstract**

Vision Transformer (ViT) has brought new breakthroughs to the field of image classification by introducing the self-attention mechanism and Graph Convolutional Networks(GCN) have been proposed and successfully applied in data representation and analysis. However, there are key challenges which limit their further development: (1) The patch size selected by ViT is crucial for accurate predictions, which raises a natural question: How to select the size of patches properly or how to comprehensively combine small patches and larger patches; (2) While the spatial structure information is important in vision tasks, the 1D position embeddings fails to capture the spatial structure information of patches more accurately; (3) The GCN can capture the local connectivity relationships between image nodes, but it lacks the ability to capture global graph structural information. On the contrary, the self-attention mechanism of ViT can draw the global relation on image patches, but it is unable to model the local structure of image. To overcome such limitations, we propose the Hierarchical Vision Transformer Enhanced by Graph Convolutional Network (GCN-HViT) for image classification. Specifically, the Hierarchical ViT we designed can model patch-wise information interactions on a global scale within each level and model hierarchical relationships between small patches and large patches across multiple levels. In addition, the proposed GCN method functions as a local feature extractor to obtain the local representation of each image patch which serves as a 2D position embedding of each patch in the 2D space. Meanwhile, it models patch-wise information interactions on a local scale within each level. Extensive experiments on 3 real-world datasets demonstrate that GCN-HViT achieves state-of-the-art performance.

## Introduction

The Vision Transformer (ViT) , as an image classification model (Dosovitskiy, et al. 2020), is based on the Transformer architecture which have dominated natural language modelling (Devlin et al. 2018; Radford et al. 2018; Brown et al. 2020; Yang et al. 2019; Peters et al. 2019; Liu et al. 2019). ViT has brought new breakthroughs to the field of image classification by introducing the self-attention mechanism (Vaswani et al. 2017) to learn global information. An image is split into fixed-size patches, and then each of them is linearly embedded. In order to make use of the order of patches in 1D sequence, the standard learnable 1D position embeddings are added to the patch embeddings., and the resulting sequence of vectors are fed to a standard Transformer encoder. The core of ViT is patch-wise self-attention, which is capable of modeling the global relationship of patches in a sequence. Recent works have explored the application of transformers to various vision tasks: image classification (Chen et al. 2020), object detection (Carion et al. 2020; Zhu et al. 2020; Zheng et al. 2020; Dai et al. 2020; Sun et al. 2020), image generation (Parmar et al. 2018), segmentation (Chen et al. 2020; Wang et al. 2020), and image enhancement (Yang et al. 2020).

Graph convolutional network (GCN), as an emerging network architecture, can effectively handle graph structure data by modeling relationship between vertexes. Many GCN models have been proposed and obtained promising performance. Among these methods, Kipf and Welling innovatively applied GCN to semi-supervised learning (Kipf and Welling et al. 2016). Qin et al. extended the original GCN to a second-order version, which took into account both spatial neighborhoods and spectral neighborhoods simultaneously (Qin et al. 2018). Wan et al. segmented the entire HSI into a set of compact image regions which can preserve the local structural information of HSI, and then a graph was constructed by treating each of the image regions as a graph node (Wan et al. 2019). Hence, graph convolution were performed on the basis of graph structure data, which was an efficient way to reduce the computational cost and improve the classification accuracy.

However, there are challenges that limit the effectiveness of these methods, which can be summarized as follows.

Firstly, the size of image patches selected by ViT, namely patch size, is crucial for accurate predictions. A small patch size contains less information and only capture local features of an image, which can lead to poor performance of the model when recognizing global structures or patterns. In addition, as the image is divided into more small patches, the model needs to process more patches, which may result in higher computational costs and longer training times. On the contrary, if the patch size is large, each patch may contain excessive information and smooth out the local features of the image, which can reduce the model's sensitivity to local textures. This raises a natural question: How to select the size of patches properly or how to comprehensively combine small patches and larger patches.

Secondly, ViT use the standard learnable 1D position embeddings which are added to the patch embeddings to retain positional information. But for vision tasks, the inputs are usually 2D images, where the pixels are highly spatially structured. While the spatial structure information is important in vision tasks, the 1D position embeddings only focus on the order of patches in 1D sequence rather than the position of patches in the 2D space. As a result, the 1D position embeddings fails to capture the spatial structure information of patches more accurately.

Last but not least, the local information aggregation mechanism enables GCN to capture the local connectivity relationships between nodes, but it lacks the ability to capture global graph structural information. For vision tasks, global graph structural information is often crucial, but GCN may not effectively utilize this information. On the contrary, the self-attention mechanism of ViT can draw the global relation on patches, but it is unable to model the image local structure.

Our key insight is that there is a natural hierarchical structure in image patches, from small patches to large patches, with information flowing from foundational elements to higher-level patterns. In addition, we propose a novel approach that integrates the advantages of GCN and ViT, which not only can model the local structure information of surrounding patches but also can obtain global representations of patches.

Motivated by the above, we propose the GCN-HViT, a Hierarchical Vision Transformer Enhanced by Graph Convolutional Network for image classification. This work follows a two-step process. We first propose the Hierarchical Vision Transformer to enhance the model's comprehension of the hierarchical structure in the 2D space. The Hierarchical ViT can identify two different hierarchical levels of patches in 2D images. Secondly, we propose the GCN method to enhance Hierarchical ViT's performance further. The GCN method dedicated to 2D images can capture the spatial structure information of patches and can replace the previous 1D positional embeddings for image classification task. The main contributions of this work are as follows:

- We analyzed the hierarchical structure in image patches. The hierarchical ViT we designed can model patch-wise information interactions on a global scale within each level and model hierarchical relationships between small patches and large patches across multiple levels.
- We develop a novel position embeddings method to capture the spatial structure information of image patches and demonstrate its advantage over the simple 1D position embeddings approach by ViT. The proposed GCN method functions as a local feature extractor to obtain the local representation of each image patch which serves as a 2D position embeddings in the 2D space. Meanwhile, it models patch-wise information interactions on a local scale within each level.
- The designed GCN-HVit is not only a comprehensive model which integrates the advantages of the two models, but also a concise and elegant model whose principles can be explained clearly. We perform experiments on 3 real-world datasets, and the experimental results demonstrate that it achieves state-of-the-art (SOTA) performance.

## Methods

As individual image patch contains less semantic information, the core of methods is to address the issue of information sparsity. To accomplish this goal, we propose the GCN-HViT. This approach mainly consists of two steps. We first propose the Hierarchical ViT to model hierarchical relationships across multiple levels by aggregating small patches into large patches in the 2D space. Secondly, we propose the GCN method to enhance Hierarchical ViT's performance further. Notably, through the GCN method, the previous 1D position embeddings can be replaced for image classification task. The overall framework of the network is shown in Figure 1.

### Hierarchical ViT

In an 2D image where the pixels are highly spatially structured, the pixels can form small patches, and small patches can form large patches, thus forming a hierarchical structure in the 2D space. Just as the visual receptive field of humans extends from the local microscopic level to the macroscopic level, we aggregate small patches into large patches in the 2D space, which overcomes the limitations of patch size and enhances the model's comprehension of the hierarchical structure in the 2D space. The Hierarchical ViT we proposed can identify two different hierarchical levels of patches in 2D images, which not only model patch-wise information interactions on a global scale within each level, but also model hierarchical relationships between small patches and large patches across multiple levels.

Specially, the image $x \in \mathbb{R}^{H \times W \times C}$ is split with no overlapping to 16 small patches and we embed each of them through the CNN whose kernel size and stride are equal to patch size.

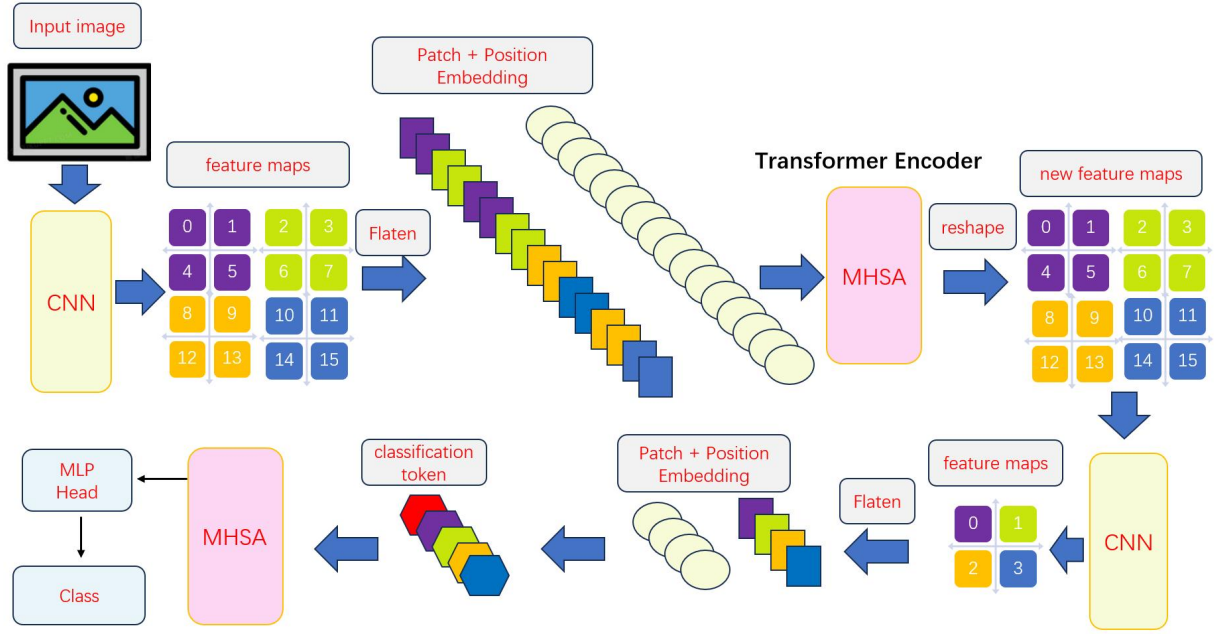

Figure 1. Model overview.

In addition, The output channels D of the CNN are the embedding dimensions D. After the feature maps produced as an output by the CNN are flattened, we obtain the input sequence of flattened 2D patches $x_p \in \mathbb{R}^{N \times D}$, which is called small patch embeddings at level 1, where $(H, W)$ is the resolution of the original image, $C$ is the number of channels, $(P, P)$ is the resolution of each image patch, and $N = HW/P^2$ is the resulting number of patches. The resulting number of patches N at level 1 is sixteen.

$$f_1 = \text{Conv}(x), \qquad (1)$$
$$x_P = \text{Flaten}(f_1), \qquad (2)$$

And then, we add position embeddings $E_{Pos1}$ obtained through the GCN method and feed the resulting sequence of vectors k to the first Transformer encoder at level 1. The Transformer encoder consists of alternating layers of multi-headed selfattention (MHSA) and MLP blocks. Layer normalization (LN) is applied before every block, and residual connections after every block.

$$k = x_P + E_{Pos1}, \; x_p \in \mathbb{R}^{N \times D}, E_{Pos1} \in \mathbb{R}^{N \times D}, \qquad (3)$$
$$k'_n = \text{MHSA}(\text{LN}(k_{n-1})) + k_{n-1}, \; n = 1 \ldots L, \qquad (4)$$
$$k_n = \text{MLP}(\text{LN}(k'_n)) + k'_n, \qquad n = 1 \ldots L, \qquad (5)$$

Notably, the output sequence of Transformer encoder at level 1 are reshaped into a new feature maps z in the 2D space through a reshape operation, which bridge the gap between level 1 and level 2.

$$z = \text{Reshape}(k_n), \qquad (6)$$

The large patch embeddings at level 2 is applied to the new feature maps z. The feature maps $z \in \mathbb{R}^{H \times W \times C}$ is split with no overlapping to 4 large patches and we embed each of them through the CNN whose kernel size and stride are equal to patch size. By aggregating small patches at level 1 into large patches at level 2, we obtain the input sequence of flattened 2D patches $z_P \in \mathbb{R}^{N \times D}$, which is called large patch embeddings at level 2. Position embeddings $E_{pos2}$ at level 2 are added as described above. The resulting number of patches N at level 2 is four.

$$f_2 = \text{Conv}(z), \qquad (7)$$
$$z_P = \text{Flaten}(f_2), \qquad (8)$$
$$h = z_P + E_{Pos2}, z_P \in \mathbb{R}^{N \times D}, E_{Pos2} \in \mathbb{R}^{N \times D}, \qquad (9)$$

An extra learnable classification token $x_{class} \in \mathbb{R}^{1 \times D}$ is also added to the sequence h and serves as the image representation. Then we feed the resulting sequence of vectors to the second Transformer encoder at level 2. A classification head CH is attached to the classification token $h_N^0$ and y is the output prediction.

$$h = [x_{class} ; h ], \ x_{class} \in \mathbb{R}^{1 \times D}, \quad (10)$$
$$h_n^{'} = MHSA(LN(h_{n-1})) + h_{n-1}, \ n = 1 \dots L, \quad (11)$$
$$h_n = MLP(LN(h_n^{'})) + h_n^{'}, \quad n = 1 \dots L, \quad (12)$$
$$y = CH(h_N^0), \quad (13)$$

**Graph Convolutional Network**

GCN specifically designed for processing graph-structured data has been widely applied in various fields in recent years. Graph data not only have nodes which can be represented as vectors, but also have information about the edges. The core idea of GCN is to perform a weighted average calculation on the information of each node's neighboring nodes as well as the information of the node itself, thereby obtaining a new feature representation. This new feature representation not only contains the information of the node itself but also integrates the information of its neighboring nodes. Therefore, it can capture the local features in graph-structured data.

In the GCN module design, patch embeddings in the 2D space can be regard as node embeddings, where each image patch represents an image node. In addition, in the 2D space, image patches or image nodes are organized in a grid-like manner and are distributed within the grid cells. Each grid cell contains an image patch. In 2D grid data, each image patch in the grid is regarded as a vertex and there is an adjacency relationship between image patches. Therefore, an adjacency matrix can be used to represent these adjacency relationships. As a result, image patches in the 2D space have adjacency relationships, which forms the adjacency matrix of image nodes. In conclusion, image patches is a type of graph data and we can apply the GCN to image patches.

The GCN receives as input a 1D sequence of patch embeddings ($x_P$ or $z_P$) and the adjacency matrices ($A_{16}$ or $A_4$). The GCN models patch-wise information interactions on a local scale and functions as a local feature extractor to obtain the local representation of each image patch. The local representation serves as a 2D position embedding $E_{Pos}$ and is added to patch embeddings, which enhances local features of each image patch. We process the input data with GCN to obtain the position embedding $E_{Pos}$.

$$E_{Pos1} = GCN(x_P, A_{16}), \quad (14)$$
$$E_{Pos2} = GCN(z_P, A_4), \quad (15)$$

# Experiments

## Datasets

For image classification, we conduct experiments on 3 real-world datasets, including Mnist, FasionMnist, and Quick Draw. The Mnist datasets consists of 70,000 grayscale images of handwritten digits (28x28 pixels), spanning 10 classes (digits 0-9). The FasionMnist datasets consists of 70,000 grayscale images of clothing items (28x28 pixels) across 10 different object classes. The Quick Draw datasets contains 20,000 grayscale images of hand-drawn doodles (64x64 pixels) across 10 different object classes. We train the models in supervised fashion. Datasets statistics are summarized in Table 1.

| Dataset | Training set | Test set | Image Size | Classes |
|---|---|---|---|---|
| Mnist | 60,000 images | 10,000 images | 28x28 pixels | 10 |
| FasionMnist | 60,000 images | 10,000 images | 28x28 pixels | 10 |
| Quick Draw | 16000 images | 4000 images | 64x64 pixels | 10 |

Table 1. Datasets statistics.

## Baselines

We first compare against the ViT, where the number of patches required for training ViT-4 is four and the number of patches required for training ViT-16 is sixteen. We further compare against Hierarchical ViT, which can identify two different hierarchical levels of patches in 2D images. Lastly, we compare GCN-HViT-2 with GCN-HViT-1.

Specially, in 2D grid data, each image patch in the grid is regarded as a vertex and there is a adjacency relationship between each image patch. Thus, the adjacency matrix can be used to represent these adjacency relationships. The GCN-HViT-1 means that we use one-way adjacency matrix. Each image patch has a one-way adjacency relationship only with image patch on its right and the image patch below it. The GCN-HViT-2 means that we use bidirectional adjacency matrix.

## Ablation study and Overall Comparison

To investigate the effectiveness of this method, we do ablation study. We first compare ViT-4 to ViT-16. ViT-16 outperforms ViT-4, which shows the importance of extracting local features from small patches. We further compare Hierarchical ViT with ViT. Hierarchical ViT outperforms with ViT, as the hierarchical relationship between small patches and large patches is crucial for enhancing the efficiency of predictions. By comparing the GCN-HViT-1 to HViT, We can observe that the GCN-HViT-1 outperforms HViT on all datasets, showing the effectiveness of introducing the GCN method. The experimental results on 3 datasets are summarized in Table 2 and Figure 2.

| Models | Mnist | FasionMnist | Quick Draw |
| --- | --- | --- | --- |
| ViT-4 | 98.57 | 89.13 | 83.65 |
| ViT-16 | 98.61 | 89.46 | 84.87 |
| HViT | 98.71 | 89.85 | 86.12 |
| GCN-HViT-2 | 98.66 | 90.03 | 86.35 |
| **GCN-HViT-1** | **98.80** | **90.26** | **86.55** |

Table 2. Summary of results in terms of classification accuracy (in percent). Specifically, GCN-HViT-2's performance is inferior to that of GCN-HViT-1, which demonstrates that the position embeddings obtained by the bidirectional adjacency matrix contains less effective information.

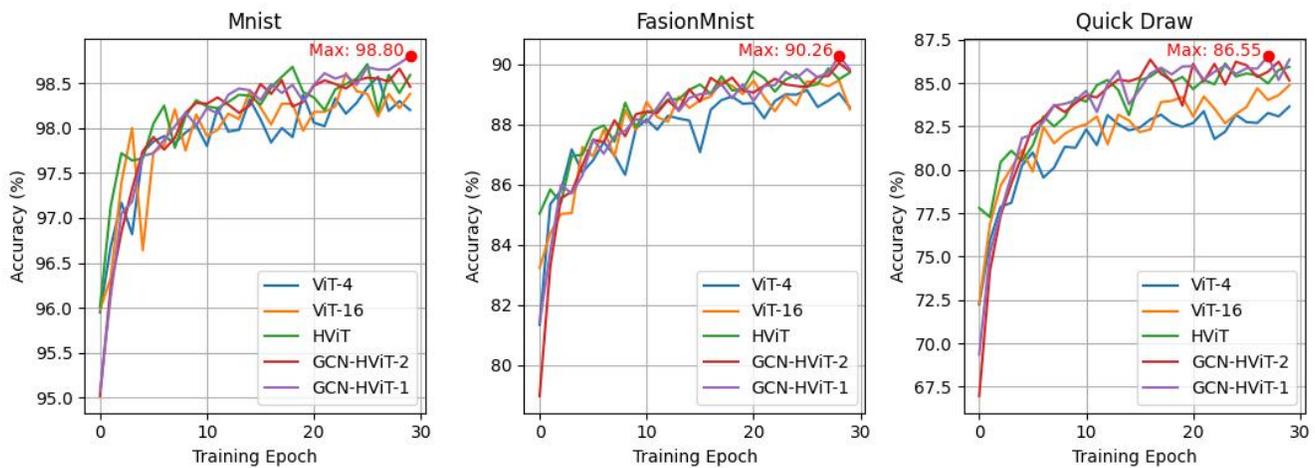

Figure 2. Test accuracy (%) plotted against training epoch for models. All models within each plot were trained on the same type of GPU to ensure a fair comparison. Notably, in our experiments, we did not perform pre-training on the models and only trained the models for 30 epochs.

## Discussion

In this paper, we propose the GCN-HViT for image classification. We analyzed the hierarchical structure in image patches, from small patches to large patches. The hierarchical transformers architecture we designed empower transformers with the ability to model patch-wise interactions on a global scale within each level and to model hierarchical relationships between small patches and large patches across multiple levels. The GCN module can model patch-wise interactions on a local scale within each level and functions as a local feature extractor to obtain the local representation of each image patch, which serves as a 2D position embeddings. Experimentally, the GCN-HViT achieves the SOTA performance.

In future work, this work can be extended in the following three aspects. First, we will apply our framework to high-resolution image datasets for image classification. Second, we will expend the two-level hierarchical framework to a multi-level one. Finally, we will extend our method to other scenarios, such as semantic segmentation and image generation.

## Reproducibility Checklist

### 1. General Paper Structure

1.1. Includes a conceptual outline and/or pseudocode description of AI methods introduced (yes/partial/no/NA) yes

1.2. Clearly delineates statements that are opinions, hypothesis, and speculation from objective facts and results (yes/no) yes

1.3. Provides well-marked pedagogical references for less-familiar readers to gain background necessary to replicate the paper (yes/no) yes

**Theoretical Contributions**

1.4. Does this paper make theoretical contributions? (yes/no) yes

  If yes, please address the following points:

1.5. All assumptions and restrictions are stated clearly and formally (yes/partial/no) yes

1.6. All novel claims are stated formally (e.g., in theorem statements) (yes/partial/no) yes

1.7. Proofs of all novel claims are included (yes/partial/no) yes

1.8. Proof sketches or intuitions are given for complex and/or novel results (yes/partial/no) yes

1.9. Appropriate citations to theoretical tools used are given (yes/partial/no) yes

1.10. All theoretical claims are demonstrated empirically to hold (yes/partial/no/NA) yes

1.11. All experimental code used to eliminate or disprove claims is included (yes/no/NA) yes

**Dataset Usage**

1.12. Does this paper rely on one or more datasets? (yes/no) yes

  If yes, please address the following points:

1.13. A motivation is given for why the experiments are conducted on the selected datasets (yes/partial/no/NA) no

1.14. All novel datasets introduced in this paper are included in a data appendix (yes/partial/no/NA) yes

1.15. All novel datasets introduced in this paper will be made publicly available upon publication of the paper with a license that allows free usage for research purposes (yes/partial/no/NA) yes

1.16. All datasets drawn from the existing literature (potentially including authors' own previously published work) are accompanied by appropriate citations (yes/no/NA) partial

1.17. All datasets drawn from the existing literature (potentially including authors' own previously published work) are publicly available (yes/partial/no/NA) yes

1.18. All datasets that are not publicly available are described in detail, with explanation why publicly available alternatives are not scientifically satisficing (yes/partial/no/NA) partial

**Computational Experiments**

1.19. Does this paper include computational experiments? (yes/no) yes

  If yes, please address the following points:

1.20. This paper states the number and range of values tried per (hyper-) parameter during development of the paper, along with the criterion used for selecting the final parameter setting (yes/partial/no/NA) no

1.21. Any code required for pre-processing data is included in the appendix (yes/partial/no) yes

1.22. All source code required for conducting and analyzing the experiments is included in a code appendix (yes/partial/no) yes

1.23. All source code required for conducting and analyzing the experiments will be made publicly available upon publication of the paper with a license that allows free usage for research purposes (yes/partial/no) yes

1.24. All source code implementing new methods have comments detailing the implementation, with references to the paper where each step comes from (yes/partial/no) no

1.25. If an algorithm depends on randomness, then the method used for setting seeds is described in a way sufficient to allow replication of results (yes/partial/no/NA) yes

1.26. This paper specifies the computing infrastructure used for running experiments (hardware and software), including GPU/CPU models; amount of memory; operating system; names and versions of relevant software libraries and frameworks (yes/partial/no) partial

1.27. This paper formally describes evaluation metrics used and explains the motivation for choosing these metrics (yes/partial/no) no

1.28. This paper states the number of algorithm runs used to compute each reported result (yes/no) no

1.29. Analysis of experiments goes beyond single-dimensional summaries of performance (e.g., average; median) to include measures of variation, confidence, or other distributional information (yes/no) no

1.30. The significance of any improvement or decrease in performance is judged using appropriate statistical tests (e.g., Wilcoxon signed-rank) (yes/partial/no) no

1.31. This paper lists all final (hyper-)parameters used for each model/algorithm in the paper's experiments (yes/partial/no/NA) partial